%% file: main.tex
\documentclass[letterpaper]{article} %
\usepackage{aaai23}  %
\usepackage{times}  %
\usepackage{helvet}  %
\usepackage{courier}  %
\usepackage[hyphens]{url}  %
\usepackage{graphicx} %
\usepackage{natbib}  %
\usepackage{caption} %
\usepackage{newfloat}
\title{Structured Case-based Reasoning for \\Inference-time Adaptation of Text-to-SQL parsers}
\author{
    Abhijeet Awasthi,
    Soumen Chakrabarti,
    Sunita Sarawagi
}
\affiliations{
    Department of Computer Science and Engineering\\

    Indian Institute of Technology Bombay, Mumbai, India\\
    \{awasthi,soumen,sunita\}@cse.iitb.ac.in
}

\usepackage{bibentry}

\usepackage{microtype}

\usepackage{amsmath}
\usepackage{defs}
\RequirePackage[inline,shortlabels]{enumitem}

\usepackage{arydshln}
\usepackage[ruled,vlined,linesnumbered]{algorithm2e}
\usepackage{adjustbox}
\usepackage{pgfplots}
\usepackage{slashbox}
\usepackage{soul}

\newcommand*\circled[1]{{\Large \textcircled{\small #1}}}

\newcommand{\roberta}{{\textsc RoBERTa}}
\newcommand{\tfive}{{\textsc T5-large}}
\newcommand{\retriever}{{\mathbf E}}
\newcommand{\sysname}{{\textsc Struct\-CBR}}
\newcommand{\seqtoseq}{{\textsc Seq2Seq}}
\newcommand{\keep}{\textsc{Keep}}
\newcommand{\ratsql}{{\textsc RAT}}
\newcommand{\smbop}{{\textsc SmBoP}}
\newcommand{\texttosql}{Text-to-SQL}
\newcommand{\beam}{B}
\newcommand{\bs}{K}
\newcommand{\z}{z}
\newcommand{\front}{F}
\newcommand{\score}{s_\theta}
\newcommand{\pscore}{p_\theta}
\newcommand{\cbrscore}{s_\phi}
\newcommand{\pcbr}{p_\phi}

\newcommand{\zb}{z_b}

\newcommand{\norm}[1]{\left\lVert#1\right\rVert}
\newcommand{\csim}{\text{sim}}

\newcommand{\transf}{\text{TX}}
\newcommand{\crossatt}{\text{XAtt}}
\newcommand{\bops}{\mathcal{B}}
\newcommand{\uops}{\mathcal{U}}

\newcommand{\zgold}{\mathcal{Z}_{\text{gold}}}
\newcommand{\cbrconcat}{ConcatCBR}
\newcommand{\knn}{GTM}
\newcommand{\model}{{\mathbf{M}}}
\newcommand{\nlqs}{{\mathcal{X}}}
\newcommand{\schemas}{{\mathcal{S}}}
\newcommand{\queries}{{\mathcal{Q}}}

\newcommand{\trainset}{{\mathcal{D}_\text{train}}}
\newcommand{\testset}{{\mathcal{D}_\text{test}}}

\newcommand{\targetdb}{\mathbf{s}_\text{new}}
\newcommand{\case}{{\mathcal{D}_\text{new}}}
\newcommand{\traincases}{C}
\newcommand{\printsql}[1]{{\footnotesize{\texttt{#1}}}}
\newcommand{\memory}{M}

\DeclareMathOperator{\logsumexp}{logsumexp}
\def\topk{\text{top-}{\bs}}
\DeclareMathOperator{\creatememory}{CreateCaseMemory}
\DeclareMathOperator{\smbopscorer}{SmBoPScores}
\DeclareMathOperator{\structcbrscorer}{StructCBRScores}
\DeclareMathOperator{\CreateTreeReps}{CreateTreeReps}
\DeclareMathOperator{\GroundTreeReps}{GroundTreeReps}
\DeclareMathOperator{\CreateJointReps}{JointReps}
\DeclareMathOperator{\TreeSimilarity}{TreeSim}
\DeclareMathOperator{\CombineScores}{CombineScores}
\DeclareMathOperator{\ted}{TED}
\DeclareMathOperator{\cosine}{CSim}

\def\expnumber#1#2{{#1}{\times}10^{#2}}

\begin{document}

\maketitle

\begin{abstract}
Inference-time adaptation methods for semantic parsing are useful for leveraging examples from newly-observed domains without repeated fine-tuning. Existing approaches typically bias the decoder by simply concatenating input-output example pairs (cases) from the new domain at the encoder’s input in a Seq-to-Seq model. Such methods cannot adequately leverage the structure of logical forms in the case examples. We propose \sysname, a structured case-based reasoning approach, which leverages subtree-level similarity between logical forms of cases and candidate outputs, resulting in better decoder decisions. For the task of adapting Text-to-SQL models to unseen schemas, we show that exploiting case examples in a structured manner via \sysname{} offers consistent performance improvements over prior inference-time adaptation methods across five different databases. To the best of our knowledge, we are the first to attempt inference-time adaptation of Text-to-SQL models, and harness trainable structured similarity between subqueries.
\end{abstract}

\section{Introduction}
Natural language interfaces to %
databases~\cite{data-geography-original,data-restaurants-logic,data-restaurants-original} %
enable access to structured information for users who are not familiar with languages like SQL by parsing user provided text-queries into executable SQLs. %
{\color{black}  Text-to-SQL semantic parsing is a challenging task that not only demands robust natural language understanding but simultaneously requires reasoning over the schema structure of the databases. Databases containing similar information (e.g. census in various countries) may be designed using diverse schema structures, thus making it hard for the model to generalize across schemas unseen during training. Hence, Text-to-SQL models often struggle to parse text queries for a new schema in a zero-shot manner~\cite{suhr2020exploring, kaggledbqa2021lee, wildtext2sql2021hazoom}. In practice, a small number of Text-to-SQL examples in the target schema are often essential for successful model adaptation. }
However, finetuning a Text-to-SQL model for each new database is not generally practical, for the following reasons:
\begin{enumerate*}[(i)]
\item  {\color{black} Huge variation in database schema makes it tedious to collect sufficiently large finetuning datasets for each schema, while finetuning on small datasets is unavoidably fraught with over-fitting, catastrophic forgetting, and instability w.r.t. random seeds.} %
\item Finetuning may take considerable time, preventing fast incorporation of new data into the model.
\item Often, a single large-footprint model serves multiple clients with diverse databases at the same time. Fine-tuning a separate model for each database is considered too resource-intensive in such multi-tenant scenarios.
\end{enumerate*} 

Therefore, we focus on fast online adaptation of {\texttosql} models without parameter updates, 
until the next cycle of finetuning is deemed feasible. 
Recently, case-based reasoning (CBR), which utilizes a memory of past labeled examples as cases, has emerged as a promising paradigm of inference-time adaptation without finetuning~\cite{das2020probcbr,cbr2021das, googlecbr2021,gupta2021retronlu}. %
CBR has been found effective for tasks like knowledge graph completion (KGC)~\cite{das2020probcbr}, question answering over knowledge bases (KBQA)~\cite{cbr2021das}, task-oriented semantic parsing~\cite{googlecbr2021, gupta2021retronlu}, translation~\cite{knnMT2021}, and text-based games~\cite{atzeni2022casebased}.  However, many prior CBR approaches designed around  {\seqtoseq} architectures simply concatenate input-output cases with the current input at the encoder~\cite{cbr2021das,googlecbr2021,gupta2021retronlu}.  These methods do not leverage the structure of logical forms (query plan trees) in case examples. \\

In response, we propose {\sysname}, a \emph{structured} CBR approach that directly exploits sub-tree level similarities between the candidate outputs and the case examples for adapting a {\texttosql} model to a new schema. 
We start with {\smbop} \citep{smbop2021rubin}, a recent semi-auto-regressive architecture that decodes query trees bottom-up, respecting the structure of SQL grammar production rules, instead of left-to-right token-level decoding in  {\seqtoseq} models~\cite{guo-etal-2019-towards, ratsql2020wang, duorat2021scholak, picardScholak2021}. We implement a novel \emph{structured case memory lookup} module to boost scores of promising candidate trees using sub-tree level similarity with case trees under similar input context. This similarity is trainable.  We show that explicitly-learned structured memory lookup leads to more accurate transfer from cases, compared to prior inference-time adaptation methods such as \cbrconcat{} and \knn~\cite{knnlm2020, cbr2021das,knnMT2021} that we implemented both on \smbop, and  other \seqtoseq\ \texttosql\ architectures like \tfive.\\

We summarize our contributions as follows:
\begin{enumerate}[1), wide, labelwidth=!, labelindent=0pt]
\item We propose  \sysname{}, which, to our knowledge, is the first inference-time adaptation method for {\texttosql} parsing without parameter fine-tuning.
\item \sysname{} incorporates a novel structured case memory and trainable query subtree similarity module that can boost scores of likely-correct outputs during inference.  This is in contrast with earlier approaches like \cbrconcat{} and \knn.
\item We propose a trainable compositional sub-tree similarity function that is both more accurate and more efficient for scoring large search frontiers, compared to default whole-tree embeddings.
\item Through experiments with five %
data\-base schemas (\S\,\ref{sec:experiments}) {\color{black} of varying complexity}, we observe that {\sysname} is consistently better than prior infer\-ence-time adap\-tation methods on both \smbop\ and sequence-based \texttosql\ models.
\item We show that {\sysname} provides almost instant 
adaptation to a target schema. In contrast, finetuning~(\S\,\ref{sec:finetuning}) can be up to 500 times slower. 
\end{enumerate}

\section{{\smbop} preliminaries}
\label{sec:background}
We present a brief background on {\smbop} here.
Readers familiar with \smbop{} can largely skip this section.
\smbop\ converts a natural language question $\bar{x} \in \nlqs$ (called the `utterance') targeting a database schema $\bar{s} \in \schemas$, to an SQL query $\hat{q} \in \queries$.  We describe the major modules in \smbop.

\paragraph{Utterance and schema encoding:}
Given token sequence $\bar{x} = [x_1, x_2, \ldots, x_n ]$ in the text query, and  database schema $\bar{s} = [s_1, s_2, \ldots, s_m]$ denoting table and column names, {\smbop} jointly encodes them
using a pre-trained Transformer like RoBERTa~\cite{liu2020roberta} followed by 
relation-aware-transformer ({\ratsql}) layers ~\cite{relationatt2018, ratsql2020wang,duorat2021scholak}. 
We denote the output from the last encoder layer as $\bar{\vx} = [\vx_1, \vx_2, \ldots \vx_n]$ and $\bar{\vs} = [\vs_1, \vs_2, \ldots \vs_m]$, representing the jointly encoded contextual embeddings of text tokens and schema elements respectively. %

\begin{figure}[t]
    \centering
    \includegraphics[scale=1.5]{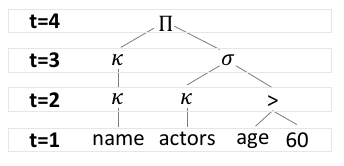}
    \caption{{\smbop}~\cite{smbop2021rubin} decodes SQL as a balanced relational algebra tree. At each level $t$, trees in the beam combine via unary or binary operators to form candidates of the next beam. {\sysname} leverages CBR on generated sub-trees.}
    \label{fig:smbop_deocding}
\end{figure}

\paragraph{Decoding SQL output:} 
Unlike standard seq\-uence-based decoding~\cite{ratsql2020wang, duorat2021scholak, picardScholak2021}, 
{\smbop} decodes the SQL tree bottom-up and in layers. \smbop\ views any SQL query as a height-balanced relational algebra tree converted using a special idempotent {\keep} operator $\kappa$ as shown 
in Figure~\ref{fig:smbop_deocding}.   
Given a beam size $\bs$, at decoding step $t$, the decoder beam $\beam_t$ comprises $\bs$ candidate sub-trees of height $t$ from the bottom.   %
At step $t+1$, trees from $\beam_t$ are grown either via unary operators (e.g. COUNT), or by combining two trees in $\beam_t$ using a binary operator (e.g. $>$), as per the SQL grammar. %
The candidate trees at step $t+1$ form a frontier set $\front_{t+1}$ and is of size $|\front_{t+1}|= \bs^2|\bops| + \bs|\uops|$, where $\bops$ and $\uops$ represent the set of binary and unary operations respectively. \smbop\ assigns each candidate tree $\z \in \front_{t+1}$ a score $\score(\z)$ (described below).
The top-$\bs$ highest scoring trees in $\front_{t+1}$ form the next beam $\beam_{t+1}$.  This continues up to a maximum height $T$, when the highest scoring tree in $\beam_T$ is output. 

\paragraph{Scoring a tree:}
A tree $z=(\zb, z_\ell, z_r)$ consists of root operator $\zb$ and subtrees $z_\ell, z_r$.
\smbop\ encodes a variable-size tree $z$ into two fixed dimensional vectors: 
\begin{enumerate*}[(i)]
\item $\vz$: an embedding of the tree computed recursively on the tree structure, where a transformer outputs $\vz=\transf_\theta([\zb, \vz_\ell,\vz_r])$; 
\item $\vz'$: a contextual representation of $\z$ grounded in input text $\bar{\vx}$ computed via a multiheaded cross attention module %
$\vz' = \crossatt_\theta(\vz, \bar{\vx})$.
\end{enumerate*}
\smbop\ computes the score of a tree $\z \in \front_{t+1}$, %
as follows:
\begin{equation}
    \score(\z
    ) = \vw_{\zb}^T \operatorname{FF}_\theta([\vz_\ell;\vz'_\ell;\vz_r;\vz'_r])
    \label{eqn:smbop_scoring_b}
\end{equation}
where $\operatorname{FF}_\theta$ is a feed forward network, and $\vw_{\zb}$ represents a learned embedding of operator $\zb$.  %

\begin{table}
\small{
\centering
\begin{tabular}{|p{0.13\linewidth} | p{0.75\linewidth}|} \hline
          Text~1 & \slshape Give the code of the airport with the fewest number of flights \\  \hline
         \smbop\ output & \printsql{SELECT sourceairport FROM flights GROUP BY sourceairport ORDER BY SUM(flightno) ASC LIMIT~1}  \\ \hline
Correct SQL & \printsql{SELECT airportcode FROM airports JOIN flights ON airportcode = sourceairport GROUP BY sourceairport ORDER BY COUNT(*) ASC LIMIT 1}\\ \hline
Text~2 & \slshape What is the code of the airport that has the highest number of flights? \\ \hline
    \end{tabular}
    \caption{Illustration of the lack of generalization of \texttosql\ to new schema.}
    \label{tab:example}
    }
\end{table}

The model is trained using Text-SQL pairs from a set of training schema to maximize the likelihood of the correct sub-trees at each beam. During inference, when presented with text utterances relating to a new database schema, the model often fails to discover the mapping of the text to schema names and relationships in the new schema. Table~\ref{tab:example} presents an example where a \smbop\ model trained on the Spider dataset~\cite{spider2018yu} is deployed on a new schema about {\tt flights}.  On inspecting the predicted and correct SQL, we find that the model failed to reason that  {\slshape number of flights} requires a {\tt count(*)} instead of {\tt sum(flightno)}.  Now suppose an expert provides the correct SQL as additional information to be used during inference of subsequent queries.  Consider a second query (shown as Text~2 in Table~\ref{tab:example}) that also needs to reason about {\slshape number of flights}, and the default \smbop\ makes similar errors (not shown). 
Only existing mechanism in \smbop\ is to fine-tune parameters which could be time-consuming and unstable.   In the next section we show how our method can instantaneously leverage test-time user labels to predict the correct SQL for Text~2. More such anecdotes appear in Table~\ref{tab:anecdotes} of the Appendix.

\section{Our proposed method: \sysname}
\label{sec:our_method}

We aim to learn a {\texttosql} model $\model$, using a dataset $\trainset$ of Text-SQL pairs such that it is capable of \textbf{C1:}~Independently translating the text queries $\bar{x}$ to executable SQL programs $\hat{q}$, and \textbf{C2:}~Utilizing a small set $\case$ of Text-SQL pairs from a target schema $\targetdb$, to improve its own predictions during inference, without finetuning. In line with prior work~\cite{das2020probcbr,cbr2021das}, we refer to the second capability \textbf{C2} as Case-based reasoning (CBR), and the dataset $\case$ of Text-SQL pairs in the target schema as cases. %

\begin{figure*}
    \centering
    \includegraphics[scale=0.9]{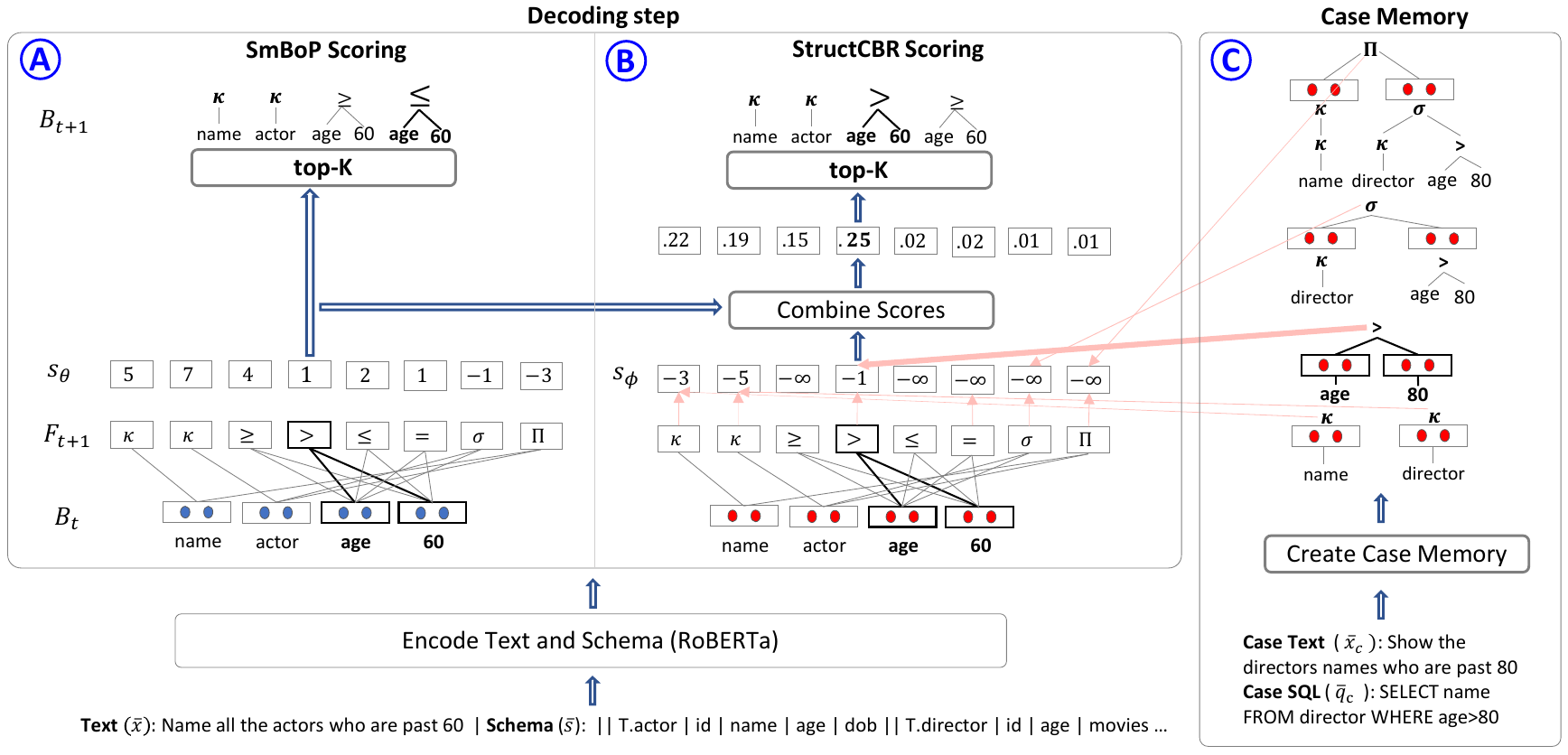}
    \caption{\textbf{Augmenting {\smbop} with {\sysname}} %
    (\underline{Struct}ured \underline{C}ase-\underline{b}ased \underline{R}easoning): In part~{\color{blue}\textbf{\circled{A}}}, the top-$\bs$ step in {\smbop} scoring misses the correct sub-tree \texttt{age > 60} due to a lower score (score=1) w.r.t. competing sub-trees in the frontier $\front_{t+1}$ like \texttt{age $\geq$ 60} (score=4) and \texttt{age $\leq$ 60} (score=2). In part~{\color{blue}\textbf{\circled{C}}}, {\sysname} creates a memory of all the sub-tree representations available in cases as described in \S~\ref{sec:case_memory}. In part~{\color{blue}\textbf{\circled{B}}}, {\sysname} scores the frontier candidates based on learned tree-similarities w.r.t.\ the sub-trees in cases as described in \S~\ref{sec:tree_sim_comp} and \S~\ref{sec:boosting_smbop}. For example, {\sysname} boosts the score of \texttt{age $>$ 60} because of its high similarity with the case sub-tree \texttt{age $>$ 80} and similarity of context {\tt who are past}. Thus, the top-$\bs$ step applied on the combined {\smbop} and {\sysname} scores recovers the correct sub-trees that otherwise may get missed based on {\smbop}'s scoring alone. %
    For brevity, we consider only one case-example in this figure.}
    \label{fig:fig1}
\end{figure*}

\label{sec:cbr-changes}
The {\sysname} module leverages the similarity %
between gold subtrees that appear in similar contexts in the set of cases $\case$ and the candidate subtrees in {\smbop}'s frontier $\front_{t+1}$, to boost the scores of likely-correct candidates at each decoding step $t+1$. 
Consider a subtree $\z$ in the frontier $\front_{t+1}$ %
for an input text $\bar{x}$, 
a case-example with text question as $\bar{x}^c$, and the gold SQL tree as  $\zgold^c$. %
Let $\z^c$ be a subtree of $\zgold^c$. The key idea of {\sysname} is, if $\z$ and $\z^c$ are structurally similar, and appear in similar contexts w.r.t. $\bar{x}$ and $\bar{x}^c$, then there is a strong evidence that the subtree $\z$ should also appear as a part of the gold tree $\zgold$ of $\bar{x}$. %
 Figure~\ref{fig:fig1} provides an illustration with $\z = \text{age} > 60$ in the candidate frontier $\front_{t+1}$, and a similarly structured case tree $\z^c = \text{age} > 80$ appearing in a similar context $\bar{x}^c$ (both contain the phrase {\tt who are past}). %

Even though the key idea of matching with case sub-trees is simple, several important design choices had to be made to ensure that CBR inter-operates efficiently with \smbop's own scoring, and  consistently improves its performance in the presence of multiple cases of varying levels of relatedness. First, how should we compute the {\em contextual} similarity of a candidate tree $z$ with a case tree, given that memory would also contain unrelated cases that would match wrong candidate trees? %
Second, how can we efficiently compute the similarity of all candidate trees with all entries in the case memory?  Unlike \seqtoseq\ models that do not perform beam-search during training, \smbop\ generates a large search frontier even during training.  We elaborate on how our design tackles these challenges next.

Algorithm~\ref{alg:highlevel} presents the high-level pseudo code, with the text in blue font representing the {\sysname} additions to the {\smbop} model. %

\begin{algorithm}[t]
\small
\textbf{input:} $\bar{x}$, $\bar{s}$, $\case$ \\
    {\color{blue} $\memory \leftarrow \creatememory(\case)$}~~~(\S~\ref{sec:case_memory}) \\
    $\bar{\vx},\bar{\vs} \leftarrow \text{EncodeTextSchema}_{\theta}(\bar{x},\bar{s})$ \\
    $\beam_{0} \leftarrow \topk \text{schema constants and DB values}$ \\ \label{line:init}
    \For{$t \leftarrow 0 \dots T-1$}{
    
        $\vz \leftarrow \CreateTreeReps(\z) $ \\
        
        $\vz' \leftarrow \GroundTreeReps(\vz,\vx)$ \\
        
        $\pscore \leftarrow  \smbopscorer(\vz, \vz')$~~~(\S~\ref{sec:background})\\
        
        {\color{blue} $G_\phi(\z,\bar{x}) \leftarrow \CreateJointReps(\vz,\vz',\vx)$}~~~(Eqn~\ref{eqn:joint_rep}) \\
        
        {\color{blue}$ \csim_\phi \leftarrow \TreeSimilarity(G_\phi(\z, \bar{x}), \memory)$}~~~(Eqn~\ref{eqn:tree_sim}) \\

        {\color{blue} $\pcbr \leftarrow \structcbrscorer(\csim_\phi, M)$}~(Eqn~\ref{eqn:cbr_score}) \\
        
        {\color{blue} $F_{t+1} \leftarrow \CombineScores(\pscore, \pcbr)$}~~~(Eqn~\ref{eqn:combined_scores}) \\

        $\beam_{t+1} \leftarrow \topk(F_{t+1})$ \label{line:prune}
 }
 \Return $\argmax_\z(\beam_T)$
\caption{{{\smbop} with {\color{blue}\sysname}.}}
\label{alg:highlevel}
\end{algorithm}

\subsection{Choosing tree representations}
\label{sec:tree_rep_comp}
We need to choose a representation of a tree $z$ using which we can efficiently compute similarity with case trees.  Just the structural similarity of $z$ with a case $z^c$ is not sufficient unless we also contextualize them on their respective inputs.  Accordingly, we design an embedding function $ G_\phi(\z, \bar{x}) \mapsto \mathbb{R}^d$ that jointly encodes a candidate tree $z$ corresponding to an input $\bar{x}$ as a $d$ dimensional vector.  We train a separate transformer model $\transf_\phi$ with parameters $\phi$ that takes as input four vectors: $\vz$ that encodes the structure of the tree $\z$,  $\vz'$ that is the contextual representation of $\z$ defined in \S~\ref{sec:background}, an embedding $\vw_b$ of $\z$'s root node $b$, and  $\text{pool}(\vx)$ a mean-pooled version of the input text representation $\bar{\vx}$:
\begin{equation}
     G_\phi(\z, \bar{x}) = \transf_\phi([\vz, \vz', \vw_b, \text{pool}(\vx)]).
     \label{eqn:joint_rep}
\end{equation}
This embedding captures both the structure and context and the parameters $\phi$ are trained to co-embed similar trees in matching contexts, while pulling apart pairs  differing either structurally or contextually. For example, in Figure~\ref{fig:fig1} if the query text was {\tt Name all actors who are 60 or above}, then the similarity of candidate {\tt age $>$ 60} from the same case sub-tree should be reduced.  
Unlike recursive tree representations \citep{socher2013reasoning}, here contextualization w.r.t. $\bar{x}$ plays a critical role.

\subsection{Case memory design}
\label{sec:case_memory}
We construct a case memory $\memory$ over the gold SQL trees $\{\zgold^c\}$ for {\em all} cases in $\case$. Corresponding to each node $b$ of a gold tree $\zgold^c$ we form a subtree rooted at $b$ and including the part of $\zgold^c$ below $b$.  Thus, the size of the case memory is the total number of nodes over all gold trees in cases. %
The encoding $G_\phi(z^c,\bar{x}^c)$ of each subtree $z^c$ for a case $(\bar{x}^c,\zgold^c)$ in $\case$ 
is pre-computed using Equation~\ref{eqn:joint_rep} and stored in $\memory$. %

\subsection{Efficient tree similarity computation}
\label{sec:tree_sim_comp}
We need to compute the similarity of each tree $\z$ in the frontier $\front_{t+1}$ %
with all case sub-trees $\z^c \in \memory$. 
One way to compute similarity between trees $\z$ and $\z^c$ is based on $\ell_2$ distance\footnote{Like~\citet{knnlm2020} we observed better results with $\ell_2$ distance, in comparison to inner product.} between their $G_\phi$ representations as follows:
\begin{equation}
\resizebox{0.85\hsize}{!}{$
\csim_\phi(\z,\z^c, \bar{x},\bar{x}^c) = - \norm{G_\phi(\z,\bar{x}) - G_\phi(\z^c,\bar{x}^c)}_2$}
\label{eqn:true_sim}
\end{equation}
However, computing $G_\phi$ representations %
for each tree $\z \in \front_{t+1}$ entails large memory and compute costs since the frontier size $|\front_{t+1}|=\bs^2|\bops| + \bs|\uops|$ is quadratic in beam-size $\bs$. With the default $\bs$ for {\smbop} being 30, and size of the \smbop\ grammar, this translates to around 23 thousand trees per frontier.  
Pruning the frontier $\front_{t+1}$ based on \smbop\ scores alone resulted in poor performance.
This led us to design an alternative compositional CBR scoring method that can more efficiently score all candidate trees in the frontier.

Our key idea is to compute the similarity between two trees compositionally as a function of similarity between their left and right children respectively. This requires only $\mathcal{O}(\bs)$ tree representations for the trees in beam $\beam_t$ %
as against $\bs^2|\bops| + \bs|\uops|$ operations of the whole-tree approach.
Recall that the trees $\z \in \front_{t+1}$ are formed by combining trees in beam $\beam_t$ via SQL operations. 
A tree $\z \in F_{t+1}$ can thus be represented as $\z = (z_b, \z_\ell, \z_r)$ where $\z_\ell$, $\z_r \in \beam_t$ denote the left and right child respectively, and $z_b$ is the root node combining $\z_\ell$ and $\z_r$.  
After the beam $\beam_t$ is created, we  compute the embedding $G_\phi$ for each tree in $\beam_t$ using Equation~\eqref{eqn:joint_rep}.
Now, the similarity %
between a candidate tree $\z = (z_b, \z_\ell, \z_r)$ %
for an input text $\bar{x}$, and a case sub-tree $\z^c = (z_b^c, \z_\ell^c, \z_r^c)$ on input text $\bar{x}^c$  in memory $\memory$ is computed as:
\begin{multline}
\widehat{\csim}_\phi(\z,\z^c,\bar{x},\bar{x}^c)
    = \csim_\phi(\z_\ell,\z^c_\ell,\bar{x},\bar{x}^c) \\ 
    + \csim_\phi(\z_r,\z^c_r,\bar{x},\bar{x}^c).
\label{eqn:tree_sim}
\end{multline}
\noindent
{\color{black}In Section~\ref{sec:experiments}, we show that using this similarity function provides better results by allowing the entire frontier to be scored more efficiently in comparison to computing similarities based on Equation~\ref{eqn:true_sim} only for a subset of trees in the frontier pruned based on {\smbop} scores.}

\subsection{Boosting {\smbop} frontier with tree similarities}
\label{sec:boosting_smbop}
To compute an overall score of a candidate tree $\z \in F_{t+1}$ based on its similarity with the case sub-trees in $\memory$, we aggregate over all the case sub-trees $\z^c$ with the same root node ($z_b=z_b^c$) using a $\logsumexp$ operator, which provides us a soft-maxed similarity of $\z$ w.r.t. case sub-trees. 
\begin{equation}
    \cbrscore(\z) = \log \hspace{-1em} \sum_{c \in \memory \land z_b=z_b^c}
    \hspace{-1em} \exp(\widehat{\csim}_\phi(\z,\z^c,\bar{x},\bar{x}^c))
    \label{eqn:cbr_score}
\end{equation}
Now every candidate tree $z  \in F_{t+1}$ has two scores: $\score(z)$ assigned by default \smbop\ and $\cbrscore(z)$ computed by \sysname.  The scores $\score(z)$ and $\cbrscore(z)$ can lie in very different ranges. Summing them in a normalized probability space provided better results than summing the scores directly. Hence, we independently normalize $\score(z)$ to $\pscore(z)$ and $\cbrscore(z)$ to $\pcbr(z)$ by a softmax operation applied over all trees in the frontier. The combined score of a frontier tree $\z$ is:
\begin{align}
p(\z) = (\pscore(\z) + \pcbr(\z))/2.
\label{eqn:combined_scores}
\end{align}

\subsection{Supervising {\sysname}}
During training, we assume availability of training data $\trainset = \{(\bar{x}_i, \bar{s}_i, \bar{q}_i)\}_{i=1}^{N}$ consisting of utterances $\bar{x}_i$ on a schema $\bar{s}_i$, and the corresponding gold SQL queries $\bar{q}_i$. We first train the {\smbop} model, parameterized as $\theta$, using $\trainset$. The training objective of \smbop\ for a single example maximizes the likelihood of sub-trees that are part of the tree $\zgold$ corresponding to gold SQL $\bar{q}$:
\begin{equation}
    \mathcal{L}_\theta = -\sum_{t=0}^T\sum_{\z_t \in \zgold}\log\pscore(\z_t).
    \label{eqn:smbop_sup}
\end{equation}
Next, we introduce the {\sysname} module parameterized as $\phi$ on top of the (now frozen)  {\smbop} model.  We observed training the {\sysname} parameters $\phi$ while freezing the learned {\smbop} parameters $\theta$ to provide slightly better results in comparison to training both $\theta$ and $\phi$ jointly. 
The parameters $\phi$ are also learned using $\trainset$ by maximizing the likelihood of the gold subtrees as per the distributions $\pcbr$ and $p$ through the following loss function: %
\begin{equation}
    \mathcal{L}_\phi = -\sum_{t=0}^T\sum_{\z_t \in \zgold}\log\pcbr(\z_t) + \log p(\z_t)
\end{equation}
The $-\log\pcbr(z_t)$ term maximizes the likelihood of gold trees w.r.t. the CBR distribution $\pcbr$, independent of the {\smbop} distribution $\pscore$. Similarly, the $-\log p(z_t)$ term maximize the likelihood of the gold trees w.r.t. the combined distribution $p$ (Eqn \ref{eqn:combined_scores}).
During training, we design each training batch to contain $\traincases$ examples from same schema so that for a given train example, the remaining $\traincases-1$ examples serve as the cases from the same schema. We train with $\traincases=32$ and a batch-size of 64.

\section{Related work}
\nocite{data-geography-original,data-restaurants-logic,data-restaurants-original}
\label{sec:related_work}

We review prior work on inference-time model adaptation %
for related tasks and also describe our adaptation of some of these works in the context of {\texttosql} for comparisons with {\sysname}.

\paragraph{Concatenating related examples with input:}
A common approach, that we call \cbrconcat, for utilizing cases during inference is to concatenate the input-output pair of each case along with the input text at the encoder of a {\seqtoseq} model. %
During training, the decoder is expected to learn to utilize the cases on the encoder side. %
\citet{cbr2021das} utilize {\cbrconcat} for question answering over knowledge bases, and  \citet{googlecbr2021, gupta2021retronlu} utilize \cbrconcat\ for other semantic parsing tasks.
{\cbrconcat} is similar to the retrieve and edit framework for structured outputs~\cite{hashimoto2018retrieve} and machine translation~\cite{hossain2020simple}.   For the {\texttosql} task, we implement a {\cbrconcat} baseline that trains an {\smbop} model to use retrieved Text-SQL examples concatenated with the input-text. During inference, the retrieval index is updated with the case-examples from the target schema.
\paragraph{Generalization through Memorization ({\knn}):}
\citet{knnlm2020,knnMT2021} propose a memory look-up based method for adapting pre-trained language and machine translation models to a target domain. Given a target dataset, their method constructs a look-up index by using contextual embeddings from the pre-trained model as keys and the corresponding text tokens as values. During inference the model scores are interpolated with the similarity scores aggregated over the nearest neighbours in the loop-up index. For our {\texttosql} set-up, we implement this baseline using a trained {\smbop} model. %
We memorize the dataset $\case$ in the target schema by creating a look-up index with embeddings of child subtrees from {\smbop} as keys: $[\vz_\ell;\vz'_\ell;\vz_r;\vz'_r]$, and their parent nodes as values.  During inference, the scores from the {\smbop} model are interpolated with neighbour similarities in a way similar to~\citet{knnMT2021}. Unlike {\sysname} and {\cbrconcat}, this baseline ({\knn}) does not explicitly train the {\smbop} model for utilizing the cases during inference.\\
We discuss other related work in Appendix~\ref{sec:add_related_work}.

\section{Experiments}
\label{sec:experiments}
We evaluate {\sysname} for adapting a {\texttosql} model to five different target schemas without finetuning. The target schemas are chosen from varying domains. 
We compare {\sysname} with prior inference-time adaptation methods %
discussed in \S~\ref{sec:related_work}, and present an ablation study. We also show that {\sysname} enables much faster adaptation of {\texttosql} models in comparison to finetuning. %

\paragraph{Datasets:}
We utilize Spider~\cite{spider2018yu}, which is a collection of {\texttosql} examples covering 200 unique schemas. We use the train split of Spider as $\trainset$,
for training all the models.  
$\trainset$ contains 7000  Text-SQL example pairs from 140 databases. For evaluation, we hold out the following five databases containing the most examples from Spider's dev set~\footnote{Spider's test set is publicly inaccessible as of 08/15/2022.}: {\{world\_1, car\_1, cre\_Doc\_Template\_Mgt, dog\_kennels, flight\_2\}}. The five evaluation databases do not occur in the train set, and belong to sufficiently different domains of varying difficulty. %
The remaining part of the dev set containing 576 examples is used for model selection while training on $\trainset$. %
We hold out 30 randomly selected examples from each of the five selected databases as $\case$ (cases) for adaptation, and use the remaining examples as the test set, $\testset$. The average size of $\testset$ is 60, and varies from roughly 50 to 90 examples across the five schemas. %
{\color{black} To ensure robust evaluation, we report numbers averaged over three random $\case$/$\testset$ splits.} %
We also report the numbers micro-averaged over all the 300 test examples across the five schemas. %

\paragraph{Evaluation metrics:}
Following prior work~\cite{spider2018yu}, we report Execution Accuracy \textbf{(EX)} and Exact-Set-Match Accuracy \textbf{(EM)} for all the methods. EX returns 1 if executing the gold query $\bar{q}$ and the predicted query $\hat{q}$ on the target database gives the same results. EM compares all the SQL clauses within $\bar{q}$ and $\hat{q}$ and returns 1 if all the clauses match, except possibly the DB-values (constants) in the SQL query. Most {\texttosql} models utilize beam search, and return the top-$\bs$ highest scoring candidates in the beam as the output. Hence, we also report the top-$\bs$ versions of EM and EX metrics as BEM and BEX respectively, where $\bs$ is the beam size. In our experiments, $\bs=30$. BEM/BEX for a beam is 1, if at least one of the candidates in the beam has an EM/EX of 1. %

\paragraph{Methods compared:}  We compare the accuracy of \sysname\ after adaptation with the following methods: \textbf{(i)}~{\smbop}: The base model without any adaptation to benchmark the gains from different inference-time adaptation methods. \textbf{(ii)}~{\cbrconcat}: The standard method of concatenating input-output case examples with the input-text. \textbf{(iii)}~{\knn}: Mapping $\case$ using {\smbop} into a non-parametric memory for augmenting model's predictions with inference-time memory look-ups similar to~\citet{knnlm2020,knnMT2021}. %
We discussed {\cbrconcat} and {\knn} baselines in Section~\ref{sec:related_work}. All the baselines are implemented using \smbop\ as the base model.  In Appendix~\ref{sec:t5_cbr} we also present \cbrconcat\ implemented on a T5-based \seqtoseq\ model.

\paragraph{Implementation details:}%
\label{sec:implementation_details}
We implemented \sysname{} and baselines using AllenNLP~\cite{Gardner2017AllenNLP} and  Transformers~\cite{hftransformers} libraries. We utilize the authors' implementation of {\smbop}~\cite{smbop2021rubin}. %
{\color{black} Due to limited computing resources, we primarily experiment with the \textsc{RoBERTa-base} checkpoint for initializing the text encoder, followed by four RAT layers~\cite{ratsql2020wang} to encode the schema structure.} %
All other hyper-parameters are the set to their default values. The {\smbop} model is trained on $\trainset$ for 60K steps with a batch size of 80, using the default learning rate (LR) of $\expnumber{1.86}{-4}$. %
The {\knn} baseline utilizes the output of this model for memory look-ups. For {\cbrconcat} baseline we train the {\smbop} model further for 60K steps with a LR of $\expnumber{5}{-5}$, while concatenating the retrieved cases in the encoder's input. {\sysname} introduces 2.53\% additional parameters ($\phi$) over the {\smbop} parameters ($\theta$). We train the parameters $\phi$ on $\trainset$ using a batch size of 64 for 60K steps with the default LR of $\expnumber{1.86}{-4}$. %
Additional training details are provided in Appendix~\ref{sec:add_hparams}. 

\paragraph{Overall Results:}
\input{spider_results_gains}
In Table~\ref{tab:spider_results}, we compare {\sysname} with different inference-time methods for adapting {\smbop} based models on five different evaluation schemas. The performance of the unadapted {\smbop} model varies from 46.1~EM to 84.3~EM  across the five schemas indicating that the evaluation schemas are of varying difficulty. We find that {\sysname} almost consistently offers substantial gains over the base {\smbop} model w.r.t. all the metrics, and across all the five schemas. {\sysname} gains upto 6.3 EM points and on average 4.6 EM points over the base {\smbop} model, while achieving almost 4 times higher gains than best existing method. 
In contrast, the {\cbrconcat} method, which has been shown to work well for other semantic parsing tasks~\cite{cbr2021das,googlecbr2021}, provides positive EM gains for only three out of five schemas, and causes overall drop in  micro-averaged EM over all the test instances. We also explored a {\cbrconcat} implementation based on {\tfive} model with a comparable base accuracy (Appendix~\ref{sec:t5_cbr}). Even  compared to this method, we continue to observe higher and more consistent gains from {\sysname} on \smbop. %
{\knn}, another inference-time adaptation baseline, utilizes memory lookups similar to~\citet{knnMT2021}, and offers almost consistent but much smaller gains %
in comparison to {\sysname}. The {\knn} baseline performs memory look-ups based on representations learned by the base {\smbop} model whereas {\sysname} is explicitly trained to perform sub-tree level memory look-ups. %
In particular, {\sysname} boosts the top-$\bs$ EM score (BEM) by up to 12.3 points. With a large-sized {\smbop} architecture we continue to observe consistent gains for most of the schemas. Section~\ref{sec:large_model} and Table~\ref{tab:spider_results_large} in Appendix provide results for adapting the large-sized models in the same setting as Table~\ref{tab:spider_results}. Table~\ref{tab:anecdotes} in Appendix provides some anecdotal examples that were originally mispredicted by the base {\smbop} model, but were fixed by {\sysname}. %

\begin{figure}[t]
    \centering
    \includegraphics[scale=0.5]{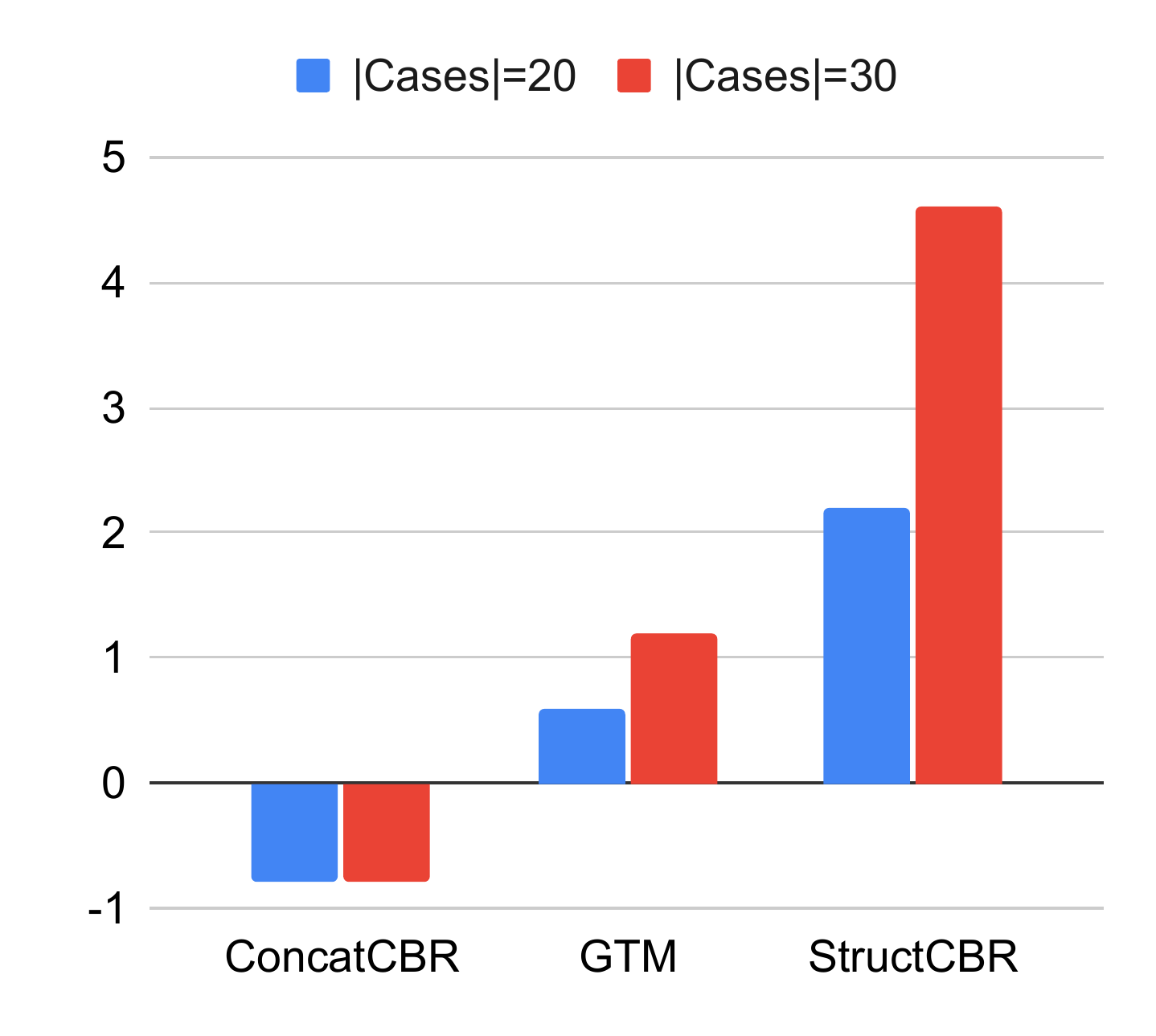}
    \caption{Impact of case size on gains in EM (y-axis) of adapted models (x-axis). With 20 cases, {\sysname} still outperforms {\knn} with 30 case examples.}
    \label{fig:case_size}
\end{figure}

\paragraph{Impact of number of cases:} We run another set of experiments by reducing the number of cases from 30 to 20. Figure~\ref{fig:case_size} shows that gains from both {\knn} and {\sysname} are smaller with fewer cases, as expected. Interestingly, {\sysname} with 20 cases still outperforms {\knn} with 30 case examples. For {\cbrconcat} we do not observe significant changes because it additionally augments the case memory with examples retrieved from Spider's train set. Not augmenting cases via retrieval resulted in even worse performance of {\cbrconcat}.

\paragraph{Justification for tree similarity:}
In Section~\ref{sec:tree_sim_comp}, we argued that directly computing tree similarities ($\csim_\phi$) using Equation~\eqref{eqn:true_sim} was inefficient, and required pruning to be practical. Instead in \sysname\ we compute tree similarity ($\widehat{\csim}_\phi$ as per  Equation~\ref{eqn:tree_sim}) more efficiently as a composition of similarity of its children, and does not require pruning.
Table~\ref{tab:just_tree_sim} justifies our design by comparing results of scoring a \emph{pruned frontier} containing top-$5\bs$ trees using $\csim_\phi$, with scoring the \emph{entire frontier} using $\widehat{\csim}_\phi$. Efficiently scoring the entire frontier provides us better results on 4 out 5 schemas and a micro-averaged gain of 2.2 points in EM.
\begin{table}[h!]
\centering
\begin{adjustbox}{max width=.95\hsize}  \tabcolsep 2pt
\begin{tabular}{|l|r|r|r|r|r||r|} \hline
     & world\_1 & car\_1 & cre & dog & flights\_2 & Micro  \\ 
      & & & \_Doc & \_kenn &  & avg  \\ \hline
Whole-tree & 43.4     & 48.9   & 88.7     & \textbf{76.2}         & 55.1 & 59.6          \\
Ours & \textbf{48.7}     & \textbf{49.4}   & \textbf{90.6}     & 70.1         & \textbf{61.2} & \textbf{61.8}   \\     \hline
\end{tabular}
\end{adjustbox}
\caption{Scoring \emph{pruned frontier} as per similarity of whole trees as in Equation~\eqref{eqn:true_sim} vs.\ our scoring of the \emph{entire frontier} as per similarity composed from subtrees, as in Equation~\eqref{eqn:tree_sim}. Better performance on four out of five schemas justifies our design of tree similarity.}
\label{tab:just_tree_sim}
\end{table}

\paragraph{Comparison with fine-tuning:}
\label{sec:finetuning}
Adapting transformer models via finetuning often provides significant accuracy improvements, thanks to recent advances in %
optimization of transformer models~\cite{pmlr-v119-huang20f,  dtfixup2021xu}. However, finetuning is not viable when a model needs to be adapted instantaneously from just a few data points, e.g. quickly incorporating expert feedback in the form of a few examples. %
In Table~\ref{tab:run_time}, we show that {\sysname} serves the purpose of instantaneously improving model performance (+2.6 pts EM), while being roughly 50$\times$ faster than finetuning for a single epoch, and 5000$\times$ faster than finetuning for 100 epochs of 30 case examples from world\_1 schema. Finetuning required atleast 10 epochs to outperform {\sysname}'s accuracy. Each epoch involved four parameter update steps of batch-size 8. We note that applying {\sysname} over {\smbop} incurs a small increase ($\sim$1.2$\times$) in inference time per example. 
Overall, we observe that on three out of five schemas {\sysname} 
instantly offers more than 50\% of gains achieved by finetuning for 100 epochs, and 43\% of gains on average across all schemas (Table~\ref{tab:relative_gains} in appendix). This establishes {\sysname} as a method for instantly utilizing available case examples for fast model adaptation until the next cycle of finetuning becomes affordable. %

\begin{table}
\centering
\adjustbox{max width=\hsize}{ \small
\begin{tabular}{|l|r|r|}
\hline
Adaptation Method   & Time(s)$\downarrow$  & EM\%$\uparrow$\\ \hline
SmBOP (Unadapted)      &  0.0   &  48.3     \\
StructCBR         &  0.1    &  50.6    \\
Finetuning (1 epochs)  &  5.0   &  48.3     \\
Finetuning (2 epochs)  &  10.0    &  47.5         \\
Finetuning (5 epochs)  &  25.0    &  48.3       \\
Finetuning (10 epochs) &  50.0    &  50.2       \\
Finetuning (20 epochs) &  100.0    &  50.9       \\
Finetuning (100 epochs) &  500.0    &  52.1       \\ \hline
\end{tabular} }
\caption{Comparing adaptation time and EM accuracy of {\sysname} and finetuning for different number of epochs on 30 cases of world\_1 schema. We report wall clock times in seconds. All the numbers were averaged over 3 runs. Finetuning took atleast 500x more time (10 epochs) to achieve EM gains that are almost instantly (0.1s) achieved by {\sysname}} 
\label{tab:run_time}
\end{table}

\section{Conclusion and Future Work}
We presented {\sysname}, a method for instant adaptation of {\texttosql} models without finetuning. We show that utilzing case examples in a more structured way via sub-tree level look-ups offers better performance in comparison to the standard method of concatenating case examples with input text into a {\seqtoseq} encoder. %
We find that explicitly learning to perform memory look-ups provides larger gains in comparison to look-ups using a pre-trained model. Finally, we show that {\sysname} enables much faster model adaptation in comparison to finetuning, potentially allowing instantaneous adaptation to expert feedback provided in form of a few examples. We propose {\sysname} as a faster alternative to model adaptation, until the next finetuning cycle is deemed feasible. 
Despite its speed, there remains an accuracy gap between \sysname{} and sufficient finetuning, which might be narrowed  by more sophisticated similarity networks.
Our exploration of {\sysname} focused only on the {\texttosql} task. In future we wish to explore the effectiveness of {\sysname} for other semantic parsing tasks.

\bibliography{custom}

\clearpage
\appendix
\section{Appendix}
\label{sec:appendix}
\setcounter{table}{0}
\renewcommand{\thetable}{A\arabic{table}}
\subsection{Details of the retriever used with {\cbrconcat}}
 The {\cbrconcat} baseline discussed in Section~\ref{sec:related_work} first retrieves Text-SQL pairs for a given utterance $\bar{x}$, which are then concatenated with the utterance $\bar{x}$ as an in input to model's encoder. Similar to~\citet{cbr2021das,poesia2022synchromesh}, we train a {\roberta}-\textsc{base} based sentence embedding model $\retriever: \nlqs \mapsto \mathrm{R}^d$, that retrieves Text-SQL pairs $\{\bar{x}_r,\bar{q}_r\}$ from cases or training data having higher cosine similarty with the input utterance $\bar{x}$. The retriever is trained independent of {\texttosql} models. For any two Text-SQL pairs $(\bar{x}_i,\bar{q}_i)$ and $(\bar{x}_j,\bar{q}_j)$ in the train set, we utilize normalized tree-edit-distance  $\ted(\bar{q}_i,\bar{q}_j) \in [0,1]$ between queries $(\bar{q}_i,\bar{q}_j)$ to supervise the cosine-similarity scores between sentence embeddings $\cosine(\retriever(\bar{x}_i), \retriever(\bar{x}_j))$, such that pairs with low tree-edit-distance have higher cosine-similarity. We utilize APTED library~\cite{pawlik2015efficient, pawlik2016tree} to compute tree-edit-distance between the relational algebra trees of the SQL queries. We modify the cost function of tree-edit-distance to ignore the leaf values and constants in the tree so that structurally similar trees are assigned lower distance. We normalize tree-edit-distance by size of the larger of two trees, so it lies in range $[0,1]$. \\
More concretely, for a given Text-SQL pair in a training batch $(\bar{x}_i,\bar{q}_i)$ we sample 15 more Text-SQL pairs $\{(\bar{x}_j,\bar{q}_j)\}_{j=1}^{15}$. Then we compute tree-edit-distances $\{\ted(\bar{q}_i,\bar{q}_j)\}_{j=1}^{15}$, and cosine similarities $\{\cosine(\retriever(\bar{x}_i), \retriever(\bar{x}_j))\}_{j=1}^{15}$ between sentences embeddings. The cosine-similarity scores are now supervised using tree-edit-distances as per loss in Equation~\ref{eqn:retriever}.
\begin{equation}
\resizebox{0.8\hsize}{!}{$
\begin{aligned}
    w_{i,j} &= \frac{\exp(1-2\ted(\bar{q}_i,\bar{q}_j))}
    {\sum_j \exp(1-2\ted(\bar{q}_i,\bar{q}_j))} \\ 
    \mathcal{L}_{\retriever} &= -\sum_{i,j}w_{i,j}\log\frac{
    \exp(\cosine(\retriever(\bar{x}_i), \retriever(\bar{x}_j)))}
    {
    \sum_j{\exp(\cosine(\retriever(\bar{x}_i), \retriever(\bar{x}_j)))}
    }
    \end{aligned}
    $}
\label{eqn:retriever}
\end{equation}
During inference on a new schema, we update the retrieval index with available cases from the new schema. Now, given a text query $\bar{x}$, we retrieve the 5 most similar examples as per cosine-similarity.

\subsection{{\cbrconcat} baseline using T5}
\label{sec:t5_cbr}
For a fair comparison, all the baselines in Section~\ref{sec:experiments} are built on the top of the {\smbop} architecture that utilizes non-auto-regressive bottom-up decoding. However, {\cbrconcat} architectures in prior works~\cite{cbr2021das,googlecbr2021} utilize the standard auto-regressive {\seqtoseq} architectures. Hence, to further ensure a fair comparison, we explored {\cbrconcat} baselines built on the top of {\tfive}~\cite{t5google2020} models. We utilize the implementation of UnifiedSKG library~\cite{xie2022unifiedskg} for {\tfive} based {\texttosql} models~\footnote{\tiny{\url{https://github.com/HKUNLP/UnifiedSKG/blob/main/configure/Salesforce/T5_large_finetune_spider_with_cell_value.cfg}}} and modify it to concatenate input-output examples at T5's encoder to get the {\cbrconcat} baseline. The {\tfive} models were intialized with an LM-Adapted ~\footnote{\tiny{\url{https://github.com/google-research/text-to-text-transfer-transformer/blob/main/released_checkpoints.md##lm-adapted-t511lm100k}}} version. The default {\tfive} based {\texttosql} model finetuned on Spider dataset has a comparable EM accuracy %
of 57.3 w.r.t. {\smbop}'s 57.2  micro-averaged across five schemas.  \\
Table~\ref{tab:t5_cbr} presents %
the EM and EX accuracies of the default {\tfive} based {\texttosql} model, and gains obtained by its {\cbrconcat} extension. Similar to Table~\ref{tab:spider_results}, gains from {\cbrconcat} are not consistent. For two out of five databases, we observe a drop in EM and EX. The micro-averaged gains across all five schemas are also small. Thus, our observations remain largely consistent with {\cbrconcat} baseline based on {\smbop}.

\begin{table}[t!]
\centering
\begin{adjustbox}{width=0.45\textwidth}
\begin{tabular}{|l|r|r|r|r|r|r|}
\hline
      & world & car & cre & dog & flights & avg  \\ 
      & \_1 & \_1 & \_Doc & \_kenn & \_2 &   \\ \hline
 \multicolumn{7}{|c|}{EM} \\ \hline  
\textsc{T5}  & 46.4     & 40.5   & 83       & 61.2         & 66        & 57.3  \\ \hline
{+\cbrconcat} & +2.3      & +1.1    & -2.5     & +10.9         & -3.4      & +1.7  \\ \hline
 \multicolumn{7}{|c|}{EX} \\ \hline  
\textsc{T5}  & 43.1     & 41.7   & 92.5       & 66.7         & 77.6        & 61  \\ \hline
{+\cbrconcat} & +4.1      & +2.8    & -8.2     & +7.5         & -5.4      & +0.7  \\ \hline
\end{tabular}
\end{adjustbox}
\caption{EM and EX performance of a {\tfive} based {\texttosql} model and gains from its {\cbrconcat} extension. Similar to Table~\ref{tab:spider_results}, we observe that {\cbrconcat} does not provide consistent gains over the unadapted model. For two schemas the gains are negative, and gains micro-averaged (avg) across five schemas are small.} %
\label{tab:t5_cbr}
\end{table}

\subsection{{\sysname} gains relative to finetuning}
Table~\ref{tab:relative_gains} shows gains of {\sysname} over {\smbop} relative to gains of finetuning over {\smbop}. $\text{Relative Gain} = \frac{+\text{StructCBR}}{+\text{Finetuning}} \times 100$. {\sysname} instantly obtains 20\% to 83\% of EM gains achieved by finetuning for 100 epochs across different schemas. For 3 out of 5 schemas relative gains are more than 50\%. Micro-averaged across all examples, {\sysname} achieves 42.8\% of gains relative to finetuning. It is important to note here that we do not position {\sysname} as an alternative to finetuning, but instead as a method of utilizing available cases for instantaneously improving model's performance until the next cycle of finetuning becomes affordable. 

\begin{table}[h!]
\centering
\begin{adjustbox}{max width=.95\hsize} 
\begin{tabular}{l|r|r|r|r|r|r}
      & world & car & cre & dog & flights & avg  \\ 
      & \_1 & \_1 & \_Doc & \_kenn & \_2 &   \\ \hline
SmBOP         & 46       & 43.3   & 84.3     & 66.6      & 55.8      & 57.2 \\
+StructCBR     & +2.6      & +6.1    & +6.3      & +3.4       & +5.5       & +4.6  \\
+Finetuning    & +6.0        & +11.7   & +7.6      & +4.8       & +27.2      & +10.7 \\ \hline
Relative Gain (\%) & 43.9     & 52.3   & 83.7     & 71.5      & 20.1      & 42.8
\end{tabular}
\end{adjustbox}
\caption{{\sysname} instantly obtains 20\% to 83\% EM gains achieved by finetuning for 100 epochs across different schemas. On 3 out 5 schemas gains are higher than 50\% of gains from finetuning. Averaged (avg) across all examples, gains are 42.8\% of finetuning. }
\label{tab:relative_gains}
\end{table}

\subsection{Additional hyperparameter and training details}
\label{sec:add_hparams}
The baseline {\smbop} model is based on {\roberta}-\text{base} architecture and contains four RAT layers. The total number of trainable parameters in this more is roughly 133M. Adding {\sysname} module on top of {\smbop} introduces only 2.53\% additional parameters. In particular, the transformer model $\transf_\phi$ used for computing $G_\phi$ representations in Equation~\ref{eqn:joint_rep} is composed of two transformer blocks of hidden\_size=256, attention\_heads=8, feedforward\_dim=1024. All the experiments were performed on a NVIDIA RTX 3060 GPU, and a NVIDIA RTX A6000. Training times for various {\texttosql} models usually varied from 12 hours to 16 hours.

\subsection{Results with a larger model size}
\label{sec:large_model}
\input{smbop_large_results_gain}

Constrained by limited computing resources for training large models, we conducted all the experiments and ablations in Section~\ref{sec:experiments} using {\smbop} models initialized with a pretrained {\roberta}-\textsc{base} checkpoint~\cite{liu2020roberta} followed by 4 RAT layers~\cite{ratsql2020wang}.
In this section, we validate our key results on a larger model size, by repeating the experiment reported in Table~\ref{tab:spider_results} of Section~\ref{sec:experiments}. More specifically, we replace {\roberta}-\textsc{base} in {\smbop} with a pretrained {\roberta}-\textsc{large} checkpoint based on grammar augmented pre-training~\cite{grappa2020yu} as used by~\citet{smbop2021rubin}. The number of RAT layers is also increased from 4 to 8. We refer to the larger {\smbop} architecture as {\smbop}-\textsc{large} (367M parameters) and the base-sized {\smbop} model used in Section~\ref{sec:experiments} as {\smbop}-\textsc{base} (133M parameters).  
Similar to Table~\ref{tab:spider_results}, Table~\ref{tab:spider_results_large} compares different inference-time methods for adapting the {\smbop}-\textsc{large} model to new schemas. The performance of {\smbop}-\textsc{base} is reported just as a reference. For each adaptation method ({\cbrconcat}, {\knn}, {\sysname}), we report it's gains over the {\smbop}-\textsc{large} model. We consistently observe positive gains from {\sysname} across all the metrics on four out of five schemas. In contrast, the gains from prior inference time adaptation methods are not consistently positive across different metrics and schemas. It is also interesting to note that the gains in top-K accuracy metrics (BEM and BEX) from {\sysname} are significantly higher than prior methods and consistently positive across all the schemas. Better performance in top-K metrics indicates that boosting candidate subtree scores via {\sysname} improves the recall of correct subtrees during beam decoding. 

\subsection{Additional Related Work}
\label{sec:add_related_work}
\paragraph{Incontext Learning with LLMs:} 
Learning new tasks via carefully designed exemplar prompts~\cite{shin2020autoprompt,le2021many,perez2021true} for large language models (LLMs) like GPT-3~\cite{gpt3brown2020language} and Codex~\cite{chen2021codex} has shown promise for many tasks~\cite{pmlr-v139-zhao21c,min2022rethinking} without requiring to finetune model parameters. For {\texttosql},
\citet{poesia2022synchromesh} retrieve and encode input-specific prompts in a way similar to the {\cbrconcat} method. They differ from {\cbrconcat} as they do not explicitly train the LLM to perform CBR with the extracted propmts. While their method achieves impressive performance by using LLMs, training significantly (100x) smaller {\texttosql} models on task-specific datasets like Spider outperforms their LLMs by considerable margins. %

\clearpage
\onecolumn
\subsection{Anecdotes}
We include some anecdotal examples where \sysname{} fixes the mistakes made by {\smbop} by utilizing case examples.
\input{anecdotes}

\end{document}

%% file: spider_results_gains.tex
\begin{table}[h!]
\centering
\begin{adjustbox}{width=0.45\textwidth}
\begin{tabular}{|l|r|r|r|r|} \hline 
          & EM & EX & BEM & BEX       \\ \hline \hline  
\multicolumn{5}{|c|}{$\targetdb=$ world\_1 \;,\; $|\testset|=89$}           \\ \hline 
SmBOP     & 46.1 & 36.3 & 55.4 & 49.0 \\ \hdashline
{\cbrconcat} & -4.1  & -4.9 & -5.2 & -6.3  \\ 
{\knn} &  +0.0   & +0.0  & +0.4 & -0.4  \\ 
StructCBR (Ours) & \textbf{+2.6}  & \textbf{+2.6} & \textbf{+3.8} & +0.0  \\  
\hline \hline

\multicolumn{5}{|c|}{$\targetdb=$ car\_1 \;,\; $|\testset|=60$}           \\ \hline 
SmBOP     & 43.3 & 45.6 & 50.6 & 53.9 \\ \hdashline
{\cbrconcat} & +3.9  & +3.9 & +4.4 & +3.4  \\ 
{\knn} & +2.2  & +2.2 & +3.9 & +2.8  \\ 
StructCBR (Ours) & \textbf{+6.1}  & \textbf{+6.7} & \textbf{+8.3} & \textbf{+7.2}   \\  
\hline \hline 
\multicolumn{5}{|c|}{$\targetdb=$ cre\_Doc\_Template\_Mgt \;,\; $|\testset|=53$}           \\ \hline 
SmBOP     & 84.3 & 89.3 & 94.3 & 96.2 \\ \hdashline
{\cbrconcat} & +5.7  & \textbf{+5.0} & +1.9 & +1.3  \\ 
{\knn} & +3.2  & +1.9 & +1.9 & +1.3  \\ 
StructCBR (Ours) & \textbf{+6.3}  & +3.8 & \textbf{+3.8} & \textbf{+2.5}  \\  
\hline \hline

\multicolumn{5}{|c|}{$\targetdb=$ dog\_kennels \;,\; $|\testset|=49$}           \\ \hline 
SmBOP     & 66.6 & 59.2 & 80.3 & 74.8 \\ \hdashline
{\cbrconcat} & +0.7  & +0.0 & -2.0 & -1.3  \\ 
{\knn} &  +1.4 & +2.7 & +0.7 & +3.4  \\ 
StructCBR (Ours) & \textbf{+3.4}  & \textbf{+6.1} & \textbf{+3.4} & \textbf{+4.1}  \\  
\hline 

\multicolumn{5}{|c|}{$\targetdb=$ flight\_2 \;,\; $|\testset|=49$}           \\ \hline 
SmBOP     & 55.8 & 67.4 & 59.2 & 80.3 \\ \hdashline
{\cbrconcat} & -8.8  & -6.1 & -2.0 & -1.3  \\ 
{\knn} & +0.0  & -1.4 & +2.1 & +0.0  \\ 
StructCBR (Ours) & \textbf{+5.5}  & \textbf{+4.1} & \textbf{+12.3} & \textbf{+4.8}  \\  
\hline \hline 

\multicolumn{5}{|c|}{Micro-Average \;,\; $|\testset|=300$}           \\ \hline 
SmBOP & 57.2 & 56.3 & 66.0 & 67.6     \\ \hdashline
{\cbrconcat} &  -0.8 & -0.8 & -1.0 &  -1.4 \\ 
{\knn} & +1.2  & +1.0 & +1.7 & +1.2  \\ 
StructCBR (Ours) &  \textbf{+4.6} & \textbf{+4.4} & \textbf{+6.0} & \textbf{+3.4}  \\  
\hline 
\end{tabular}
\end{adjustbox}
\caption{Comparison of {\sysname} with prior inference time adaptation methods ({\cbrconcat} and {\knn}) on 5 different schemas. {\smbop} row provides the performance of the unadapted model, while other rows report gains over {\smbop} after adaptation. 
Micro-average refers to numbers averaged over all the test instances spanning across the five schemas. $|\testset|$ refers to size of the test-set, and $\targetdb$ provides the schema name.} 
\label{tab:spider_results}
\end{table}

%% file: smbop_large_results_gain.tex
\begin{table}[t]
\centering
\begin{adjustbox}{width=0.45\textwidth}
\begin{tabular}{|l|r|r|r|r|} \hline 
          & EM & EX & BEM & BEX       \\ \hline \hline  
\multicolumn{5}{|c|}{$\targetdb=$ world\_1 \;,\; $|\testset|=89$}           \\ \hline 

{\smbop}-\textsc{base}     & 46.1 & 36.3 & 55.4 & 49.0 \\ 
{\smbop}-\textsc{large}     & 50.6 & 42.3 & 58.1 & 49.8 \\ \hdashline
{\cbrconcat} & -2.3  & -0.4 & -1.9 & +0.7  \\ 
{\knn} &  -0.8   & -0.8  & +0.7 & +1.1  \\ 
StructCBR (Ours) & -1.5  & -0.7 & \textbf{+3.4} & \textbf{+2.2}  \\  
\hline \hline 

\multicolumn{5}{|c|}{$\targetdb=$ car\_1 \;,\; $|\testset|=60$}           \\ \hline 
{\smbop}-\textsc{base}     & 43.3 & 45.6 & 50.6 & 53.9 \\ 
{\smbop}-\textsc{large}     & 53.3 & 57.2 & 66.7 & 67.8 \\ \hdashline
{\cbrconcat} & +3.9  & -6.1 & -2.2 & -2.8  \\ 
{\knn} & \textbf{+6.6}  & \textbf{+3.3} & \textbf{+3.3} & -0.5  \\ 
StructCBR (Ours) & +3.3  & \textbf{+3.3} & +1.1 & \textbf{+1.7}   \\  
\hline \hline 

\multicolumn{5}{|c|}{$\targetdb=$ cre\_Doc\_Template\_Mgt \;,\; $|\testset|=53$}           \\ \hline 
{\smbop}-\textsc{base}     & 84.3 & 89.3 & 94.3 & 96.2 \\ 
{\smbop}-\textsc{large}     & 90.6 & 95.6 & 96.2 & 98.1 \\ \hdashline
{\cbrconcat} & \textbf{+3.7}  & +0.6 & \textbf{+3.2} & +1.3  \\ 
{\knn} & -1.9  & -1.2 & +1.3 & +0.6  \\ 
StructCBR (Ours) & +3.1  & \textbf{+1.9} & \textbf{+3.2} & \textbf{+1.9}  \\  
\hline \hline 

\multicolumn{5}{|c|}{$\targetdb=$ dog\_kennels \;,\; $|\testset|=49$}           \\ \hline 
{\smbop}-\textsc{base}     & 66.6 & 59.2 & 80.3 & 74.8 \\ 
{\smbop}-\textsc{large}     & 70.1 & 62.6 & 81.0 & 74.1 \\ \hdashline
{\cbrconcat} & +1.3  & +2.7 & +2.7 & +6.2  \\ 
{\knn} &  -0.7 & -0.7 & -1.4 & +4.8  \\ 
StructCBR (Ours) & \textbf{+2.1}  & \textbf{+3.4} & \textbf{+5.4} & \textbf{+8.2}  \\  
\hline \hline

\multicolumn{5}{|c|}{$\targetdb=$ flight\_2 \;,\; $|\testset|=49$}           \\ \hline 
{\smbop}-\textsc{base}     & 55.8 & 67.4 & 59.2 & 80.3 \\ 
{\smbop}-\textsc{large}     & 59.9 & 65.3 & 66.0 & 75.5 \\ \hdashline
{\cbrconcat} & \textbf{+6.1}  & \textbf{+10.9} & +5.4 & +11.6  \\ 
{\knn} & +1.3  & +0.7 & +4.7 & +7.5  \\ 
StructCBR (Ours) & +4.1  & +3.4 & \textbf{+14.9} & \textbf{+12.2}  \\  
\hline \hline 

\multicolumn{5}{|c|}{\textbf{Micro-Average} \;,\; $|\testset|=300$}           \\ \hline 
{\smbop}-\textsc{base} & 57.2 & 56.3 & 66.0 & 67.6     \\ 
{\smbop}-\textsc{large} & 62.6 & 61.8 & 71.3 & 70.1     \\ \hdashline
{\cbrconcat} &  \textbf{+2.3} & +1.0 & +1.1 &  +2.8 \\ 
{\knn} & +1.2  & +0.2 & +1.9 & +2.3  \\ 
StructCBR (Ours) &  +2.1 & \textbf{+1.9} & \textbf{+5.3} & \textbf{+4.7}  \\  
\hline 
\end{tabular}
\end{adjustbox}
\caption{Results using {\smbop}-\textsc{large}:
We compare {\sysname} with prior inference time adaptation methods ({\cbrconcat} and {\knn}) for adapting {\smbop}-\textsc{large} on 5 different schemas. {\smbop}-\textsc{base} and {\smbop}-\textsc{large} rows provide the performance of unadapted model {\smbop} models, while other rows report gains over {\smbop}-\textsc{large} after adaptation. {\smbop}-\textsc{base} numbers are just of reference. 
Micro-average refers to numbers averaged over all the test instances spanning across the five schemas. $|\testset|$ refers to size of the test-set, and $\targetdb$ provides the schema name.} 
\label{tab:spider_results_large}
\end{table}

%% file: anecdotes.tex
\begin{table*}[h!]
\small
\centering
\begin{tabular}{p{0.15\linewidth} | p{0.85\linewidth}}

Text & What is the code of airport that has the highest number of flights? \\
Incorrect SQL & \printsql{SELECT \textbf{flights.sourceairport} FROM flights GROUP BY flights.sourceairport ORDER BY SUM( flights.flightno ) DESC LIMIT 1}  \\
Corrected SQL & \printsql{\textbf{SELECT airports.airportcode} FROM \textbf{airports JOIN flights} ON airports.airportcode = flights.sourceairport GROUP BY flights.sourceairport ORDER BY COUNT( * ) DESC LIMIT 1}\\
Case Text & Give the code of the airport with the least flights.\\
Case SQL & \printsql{\textbf{SELECT airports.airportcode} FROM \textbf{airports JOIN flights} ON airports.airportcode = flights.sourceairport GROUP BY flights.sourceairport ORDER BY COUNT( * ) ASC LIMIT 1}\\ 
Mistake(s) Fixed & Correct column selection, Add missing join, Replace \printsql{SUM} by \printsql{COUNT}  \\ \hline \hline

Text &  Which airlines have a flight with source airport AHD?\\
Incorrect SQL & \printsql{SELECT flights.airline FROM  \textbf{airlines JOIN flights ON airlines.uid = flights.airline JOIN airports ON flights.sourceairport = airports.airportcode} WHERE airports.airportname = 'AHD'}  \\
Corrected SQL & \printsql{SELECT airlines.airline FROM \textbf{airlines JOIN flights ON airlines.uid = flights.airline} WHERE flights.sourceairport = 'AHD'}\\
Case Text & What are airlines that have flights arriving at airport 'AHD'?\\
Case SQL & \printsql{SELECT airlines.airline FROM \textbf{airlines JOIN flights ON airlines.uid = flights.airline} WHERE flights.destairport = 'AHD'}\\ 
Mistake(s) Fixed & Drop additional condition on Join \\ \hline \hline

Text &  What are the population and life expectancies in Brazil?\\
Incorrect SQL & \printsql{SELECT country.population , country.lifeexpectancy \textbf{FROM city JOIN country ON city.countrycode = country.code} WHERE country.name = 'Brazil'}  \\
Corrected SQL & \printsql{SELECT country.lifeexpectancy , country.population \textbf{FROM country} WHERE country.name = 'Brazil'}\\
Case Text & What are the region and population of Angola?\\
Case SQL & \printsql{SELECT Population ,  Region \textbf{FROM country} WHERE Name  =  'Angola'} \\
Mistake(s) Fixed & Drop the join condition  \\ \hline \hline

Text & Return the codes of countries that do not speak English and do not have Republics for governments. \\
Incorrect SQL & \printsql{SELECT country.code FROM country WHERE countrylanguage.language = 'English' \textbf{INTERSECT} SELECT countrylanguage.countrycode FROM country WHERE country.governmentform != 'Republic'}  \\
Corrected SQL & \printsql{SELECT country.code FROM country WHERE country.governmentform != 'Republic' \textbf{EXCEPT} SELECT countrylanguage.countrycode FROM countrylanguage WHERE countrylanguage.language = 'English'}\\
Case Text & What are the codes of the countries that do not speak English and whose government forms are not Republic?\\
Case SQL & \printsql{SELECT country.code FROM country WHERE country.governmentform != 'Republic' \textbf{EXCEPT} SELECT countrylanguage.countrycode FROM countrylanguage WHERE countrylanguage.language = 'English'}\\ 
Mistake(s) Fixed & Set operation and Equality \\ \hline \hline

\end{tabular}
\caption{Anecdotal examples where mistakes in output SQLs generated by {\smbop} are fixed with help of {\sysname} through related examples in case memory. We show only one of the related examples from cases for brevity}
\label{tab:anecdotes}
\end{table*}

%% file: main.bbl
\begin{thebibliography}{41}
\providecommand{\natexlab}[1]{#1}

\bibitem[{Atzeni et~al.(2022)Atzeni, Dhuliawala, Murugesan, and
  SACHAN}]{atzeni2022casebased}
Atzeni, M.; Dhuliawala, S.~Z.; Murugesan, K.; and SACHAN, M. 2022.
\newblock Case-based Reasoning for Better Generalization in Text-Adventure
  Games.
\newblock In \emph{International Conference on Learning Representations}.

\bibitem[{Brown et~al.(2020)Brown, Mann, Ryder, Subbiah, Kaplan, Dhariwal,
  Neelakantan, Shyam, Sastry, Askell et~al.}]{gpt3brown2020language}
Brown, T.; Mann, B.; Ryder, N.; Subbiah, M.; Kaplan, J.~D.; Dhariwal, P.;
  Neelakantan, A.; Shyam, P.; Sastry, G.; Askell, A.; et~al. 2020.
\newblock Language models are few-shot learners.
\newblock \emph{Advances in neural information processing systems}, 33:
  1877--1901.

\bibitem[{Chen et~al.(2021)Chen, Tworek, Jun, Yuan, de~Oliveira~Pinto, Kaplan,
  Edwards, Burda, Joseph, Brockman, Ray, Puri, Krueger, Petrov, Khlaaf, Sastry,
  Mishkin, Chan, Gray, Ryder, Pavlov, Power, Kaiser, Bavarian, Winter, Tillet,
  Such, Cummings, Plappert, Chantzis, Barnes, Herbert-Voss, Guss, Nichol,
  Paino, Tezak, Tang, Babuschkin, Balaji, Jain, Saunders, Hesse, Carr, Leike,
  Achiam, Misra, Morikawa, Radford, Knight, Brundage, Murati, Mayer, Welinder,
  McGrew, Amodei, McCandlish, Sutskever, and Zaremba}]{chen2021codex}
Chen, M.; Tworek, J.; Jun, H.; Yuan, Q.; de~Oliveira~Pinto, H.~P.; Kaplan, J.;
  Edwards, H.; Burda, Y.; Joseph, N.; Brockman, G.; Ray, A.; Puri, R.; Krueger,
  G.; Petrov, M.; Khlaaf, H.; Sastry, G.; Mishkin, P.; Chan, B.; Gray, S.;
  Ryder, N.; Pavlov, M.; Power, A.; Kaiser, L.; Bavarian, M.; Winter, C.;
  Tillet, P.; Such, F.~P.; Cummings, D.; Plappert, M.; Chantzis, F.; Barnes,
  E.; Herbert-Voss, A.; Guss, W.~H.; Nichol, A.; Paino, A.; Tezak, N.; Tang,
  J.; Babuschkin, I.; Balaji, S.; Jain, S.; Saunders, W.; Hesse, C.; Carr,
  A.~N.; Leike, J.; Achiam, J.; Misra, V.; Morikawa, E.; Radford, A.; Knight,
  M.; Brundage, M.; Murati, M.; Mayer, K.; Welinder, P.; McGrew, B.; Amodei,
  D.; McCandlish, S.; Sutskever, I.; and Zaremba, W. 2021.
\newblock Evaluating Large Language Models Trained on Code.

\bibitem[{Das et~al.(2020)Das, Godbole, Monath, Zaheer, and
  McCallum}]{das2020probcbr}
Das, R.; Godbole, A.; Monath, N.; Zaheer, M.; and McCallum, A. 2020.
\newblock Probabilistic Case-based Reasoning for Open-World Knowledge Graph
  Completion.
\newblock In \emph{Findings of the Association for Computational Linguistics:
  EMNLP 2020}, 4752--4765.

\bibitem[{Das et~al.(2021)Das, Zaheer, Thai, Godbole, Perez, Lee, Tan,
  Polymenakos, and McCallum}]{cbr2021das}
Das, R.; Zaheer, M.; Thai, D.; Godbole, A.; Perez, E.; Lee, J.-Y.; Tan, L.;
  Polymenakos, L.; and McCallum, A. 2021.
\newblock Case-based Reasoning for Natural Language Queries over Knowledge
  Bases.
\newblock \emph{arXiv preprint arXiv:2104.08762}.

\bibitem[{Gardner et~al.(2017)Gardner, Grus, Neumann, Tafjord, Dasigi, Liu,
  Peters, Schmitz, and Zettlemoyer}]{Gardner2017AllenNLP}
Gardner, M.; Grus, J.; Neumann, M.; Tafjord, O.; Dasigi, P.; Liu, N.~F.;
  Peters, M.; Schmitz, M.; and Zettlemoyer, L.~S. 2017.
\newblock AllenNLP: A Deep Semantic Natural Language Processing Platform.

\bibitem[{Guo et~al.(2019)Guo, Zhan, Gao, Xiao, Lou, Liu, and
  Zhang}]{guo-etal-2019-towards}
Guo, J.; Zhan, Z.; Gao, Y.; Xiao, Y.; Lou, J.-G.; Liu, T.; and Zhang, D. 2019.
\newblock Towards Complex Text-to-{SQL} in Cross-Domain Database with
  Intermediate Representation.
\newblock In \emph{Proceedings of the 57th Annual Meeting of the Association
  for Computational Linguistics}, 4524--4535. Florence, Italy: Association for
  Computational Linguistics.

\bibitem[{Gupta et~al.(2021)Gupta, Shrivastava, Sagar, Aghajanyan, and
  Savenkov}]{gupta2021retronlu}
Gupta, V.; Shrivastava, A.; Sagar, A.; Aghajanyan, A.; and Savenkov, D. 2021.
\newblock RETRONLU: Retrieval Augmented Task-Oriented Semantic Parsing.
\newblock \emph{arXiv preprint arXiv:2109.10410}.

\bibitem[{Hashimoto et~al.(2018)Hashimoto, Guu, Oren, and
  Liang}]{hashimoto2018retrieve}
Hashimoto, T.~B.; Guu, K.; Oren, Y.; and Liang, P.~S. 2018.
\newblock A retrieve-and-edit framework for predicting structured outputs.
\newblock \emph{Advances in Neural Information Processing Systems}, 31.

\bibitem[{Hazoom, Malik, and Bogin(2021)}]{wildtext2sql2021hazoom}
Hazoom, M.; Malik, V.; and Bogin, B. 2021.
\newblock Text-to-{SQL} in the Wild: A Naturally-Occurring Dataset Based on
  Stack Exchange Data.
\newblock In \emph{Proceedings of the 1st Workshop on Natural Language
  Processing for Programming (NLP4Prog 2021)}, 77--87. Online: Association for
  Computational Linguistics.

\bibitem[{Hossain, Ghazvininejad, and Zettlemoyer(2020)}]{hossain2020simple}
Hossain, N.; Ghazvininejad, M.; and Zettlemoyer, L. 2020.
\newblock Simple and effective retrieve-edit-rerank text generation.
\newblock In \emph{Proceedings of the 58th Annual Meeting of the Association
  for Computational Linguistics}, 2532--2538.

\bibitem[{Huang et~al.(2020)Huang, Perez, Ba, and Volkovs}]{pmlr-v119-huang20f}
Huang, X.~S.; Perez, F.; Ba, J.; and Volkovs, M. 2020.
\newblock Improving Transformer Optimization Through Better Initialization.
\newblock In III, H.~D.; and Singh, A., eds., \emph{Proceedings of the 37th
  International Conference on Machine Learning}, volume 119 of
  \emph{Proceedings of Machine Learning Research}, 4475--4483. PMLR.

\bibitem[{Khandelwal et~al.(2021)Khandelwal, Fan, Jurafsky, Zettlemoyer, and
  Lewis}]{knnMT2021}
Khandelwal, U.; Fan, A.; Jurafsky, D.; Zettlemoyer, L.; and Lewis, M. 2021.
\newblock Nearest Neighbor Machine Translation.
\newblock In \emph{International Conference on Learning Representations}.

\bibitem[{Khandelwal et~al.(2020)Khandelwal, Levy, Jurafsky, Zettlemoyer, and
  Lewis}]{knnlm2020}
Khandelwal, U.; Levy, O.; Jurafsky, D.; Zettlemoyer, L.; and Lewis, M. 2020.
\newblock Generalization through Memorization: Nearest Neighbor Language
  Models.
\newblock In \emph{International Conference on Learning Representations}.

\bibitem[{Le~Scao and Rush(2021)}]{le2021many}
Le~Scao, T.; and Rush, A.~M. 2021.
\newblock How many data points is a prompt worth?
\newblock In \emph{Proceedings of the 2021 Conference of the North American
  Chapter of the Association for Computational Linguistics: Human Language
  Technologies}, 2627--2636.

\bibitem[{Lee, Polozov, and Richardson(2021)}]{kaggledbqa2021lee}
Lee, C.-H.; Polozov, O.; and Richardson, M. 2021.
\newblock {K}aggle{DBQA}: Realistic Evaluation of Text-to-{SQL} Parsers.
\newblock In \emph{Proceedings of the 59th Annual Meeting of the Association
  for Computational Linguistics and the 11th International Joint Conference on
  Natural Language Processing (Volume 1: Long Papers)}, 2261--2273. Online:
  Association for Computational Linguistics.

\bibitem[{Liu et~al.(2020)Liu, Ott, Goyal, Du, Joshi, Chen, Levy, Lewis,
  Zettlemoyer, and Stoyanov}]{liu2020roberta}
Liu, Y.; Ott, M.; Goyal, N.; Du, J.; Joshi, M.; Chen, D.; Levy, O.; Lewis, M.;
  Zettlemoyer, L.; and Stoyanov, V. 2020.
\newblock Ro{\{}BERT{\}}a: A Robustly Optimized {\{}BERT{\}} Pretraining
  Approach.

\bibitem[{Min et~al.(2022)Min, Lyu, Holtzman, Artetxe, Lewis, Hajishirzi, and
  Zettlemoyer}]{min2022rethinking}
Min, S.; Lyu, X.; Holtzman, A.; Artetxe, M.; Lewis, M.; Hajishirzi, H.; and
  Zettlemoyer, L. 2022.
\newblock Rethinking the Role of Demonstrations: What Makes In-Context Learning
  Work?
\newblock \emph{arXiv preprint arXiv:2202.12837}.

\bibitem[{Pasupat, Zhang, and Guu(2021)}]{googlecbr2021}
Pasupat, P.; Zhang, Y.; and Guu, K. 2021.
\newblock Controllable Semantic Parsing via Retrieval Augmentation.
\newblock In \emph{Proceedings of the 2021 Conference on Empirical Methods in
  Natural Language Processing}, 7683--7698.

\bibitem[{Pawlik and Augsten(2015)}]{pawlik2015efficient}
Pawlik, M.; and Augsten, N. 2015.
\newblock Efficient computation of the tree edit distance.
\newblock \emph{ACM Transactions on Database Systems (TODS)}, 40(1): 1--40.

\bibitem[{Pawlik and Augsten(2016)}]{pawlik2016tree}
Pawlik, M.; and Augsten, N. 2016.
\newblock Tree edit distance: Robust and memory-efficient.
\newblock \emph{Information Systems}, 56: 157--173.

\bibitem[{Perez, Kiela, and Cho(2021)}]{perez2021true}
Perez, E.; Kiela, D.; and Cho, K. 2021.
\newblock True few-shot learning with language models.
\newblock \emph{Advances in Neural Information Processing Systems}, 34.

\bibitem[{Poesia et~al.(2022)Poesia, Polozov, Le, Tiwari, Soares, Meek, and
  Gulwani}]{poesia2022synchromesh}
Poesia, G.; Polozov, O.; Le, V.; Tiwari, A.; Soares, G.; Meek, C.; and Gulwani,
  S. 2022.
\newblock Synchromesh: Reliable code generation from pre-trained language
  models.
\newblock \emph{arXiv preprint arXiv:2201.11227}.

\bibitem[{Popescu, Etzioni, and Kautz(2003)}]{data-restaurants-original}
Popescu, A.-M.; Etzioni, O.; and Kautz, H. 2003.
\newblock Towards a Theory of Natural Language Interfaces to Databases.
\newblock In \emph{Proceedings of the 8th International Conference on
  Intelligent User Interfaces}, 149--157.

\bibitem[{Raffel et~al.(2020)Raffel, Shazeer, Roberts, Lee, Narang, Matena,
  Zhou, Li, and Liu}]{t5google2020}
Raffel, C.; Shazeer, N.; Roberts, A.; Lee, K.; Narang, S.; Matena, M.; Zhou,
  Y.; Li, W.; and Liu, P.~J. 2020.
\newblock Exploring the Limits of Transfer Learning with a Unified Text-to-Text
  Transformer.
\newblock \emph{Journal of Machine Learning Research}, 21(140): 1--67.

\bibitem[{Rubin and Berant(2021)}]{smbop2021rubin}
Rubin, O.; and Berant, J. 2021.
\newblock SmBoP: Semi-autoregressive Bottom-up Semantic Parsing.
\newblock In \emph{Proceedings of the 2021 Conference of the North American
  Chapter of the Association for Computational Linguistics: Human Language
  Technologies}, 311--324.

\bibitem[{Scholak et~al.(2021)Scholak, Li, Bahdanau, de~Vries, and
  Pal}]{duorat2021scholak}
Scholak, T.; Li, R.; Bahdanau, D.; de~Vries, H.; and Pal, C. 2021.
\newblock DuoRAT: Towards Simpler Text-to-SQL Models.
\newblock In \emph{Proceedings of the 2021 Conference of the North American
  Chapter of the Association for Computational Linguistics: Human Language
  Technologies}, 1313--1321.

\bibitem[{Scholak, Schucher, and Bahdanau(2021)}]{picardScholak2021}
Scholak, T.; Schucher, N.; and Bahdanau, D. 2021.
\newblock PICARD - Parsing Incrementally for Constrained Auto-Regressive
  Decoding from Language Models.
\newblock In \emph{Proceedings of the 2021 Conference on Empirical Methods in
  Natural Language Processing}. Association for Computational Linguistics.

\bibitem[{Shaw, Uszkoreit, and Vaswani(2018)}]{relationatt2018}
Shaw, P.; Uszkoreit, J.; and Vaswani, A. 2018.
\newblock Self-Attention with Relative Position Representations.
\newblock In \emph{Proceedings of the 2018 Conference of the North {A}merican
  Chapter of the Association for Computational Linguistics: Human Language
  Technologies, Volume 2 (Short Papers)}, 464--468. New Orleans, Louisiana:
  Association for Computational Linguistics.

\bibitem[{Shin et~al.(2020)Shin, Razeghi, Logan~IV, Wallace, and
  Singh}]{shin2020autoprompt}
Shin, T.; Razeghi, Y.; Logan~IV, R.~L.; Wallace, E.; and Singh, S. 2020.
\newblock AutoPrompt: Eliciting Knowledge from Language Models with
  Automatically Generated Prompts.
\newblock In \emph{Proceedings of the 2020 Conference on Empirical Methods in
  Natural Language Processing (EMNLP)}, 4222--4235.

\bibitem[{Socher et~al.(2013)Socher, Chen, Manning, and
  Ng}]{socher2013reasoning}
Socher, R.; Chen, D.; Manning, C.~D.; and Ng, A. 2013.
\newblock Reasoning with neural tensor networks for knowledge base completion.
\newblock \emph{Advances in neural information processing systems}, 26.

\bibitem[{Suhr et~al.(2020)Suhr, Chang, Shaw, and Lee}]{suhr2020exploring}
Suhr, A.; Chang, M.-W.; Shaw, P.; and Lee, K. 2020.
\newblock Exploring unexplored generalization challenges for cross-database
  semantic parsing.
\newblock In \emph{Proceedings of the 58th Annual Meeting of the Association
  for Computational Linguistics}, 8372--8388.

\bibitem[{Tang and Mooney(2000)}]{data-restaurants-logic}
Tang, L.~R.; and Mooney, R.~J. 2000.
\newblock Automated Construction of Database Interfaces: Intergrating
  Statistical and Relational Learning for Semantic Parsing.
\newblock In \emph{2000 Joint {SIGDAT} Conference on Empirical Methods in
  Natural Language Processing and Very Large Corpora}, 133--141. Hong Kong,
  China: Association for Computational Linguistics.

\bibitem[{Wang et~al.(2020)Wang, Shin, Liu, Polozov, and
  Richardson}]{ratsql2020wang}
Wang, B.; Shin, R.; Liu, X.; Polozov, O.; and Richardson, M. 2020.
\newblock RAT-SQL: Relation-Aware Schema Encoding and Linking for Text-to-SQL
  Parsers.
\newblock In \emph{Proceedings of the 58th Annual Meeting of the Association
  for Computational Linguistics}, 7567--7578.

\bibitem[{Wolf et~al.(2020)Wolf, Debut, Sanh, Chaumond, Delangue, Moi, Cistac,
  Rault, Louf, Funtowicz, Davison, Shleifer, von Platen, Ma, Jernite, Plu, Xu,
  Scao, Gugger, Drame, Lhoest, and Rush}]{hftransformers}
Wolf, T.; Debut, L.; Sanh, V.; Chaumond, J.; Delangue, C.; Moi, A.; Cistac, P.;
  Rault, T.; Louf, R.; Funtowicz, M.; Davison, J.; Shleifer, S.; von Platen,
  P.; Ma, C.; Jernite, Y.; Plu, J.; Xu, C.; Scao, T.~L.; Gugger, S.; Drame, M.;
  Lhoest, Q.; and Rush, A.~M. 2020.
\newblock Transformers: State-of-the-Art Natural Language Processing.
\newblock In \emph{Proceedings of the 2020 Conference on Empirical Methods in
  Natural Language Processing: System Demonstrations}, 38--45. Online:
  Association for Computational Linguistics.

\bibitem[{Xie et~al.(2022)Xie, Wu, Shi, Zhong, Scholak, Yasunaga, Wu, Zhong,
  Yin, Wang et~al.}]{xie2022unifiedskg}
Xie, T.; Wu, C.~H.; Shi, P.; Zhong, R.; Scholak, T.; Yasunaga, M.; Wu, C.-S.;
  Zhong, M.; Yin, P.; Wang, S.~I.; et~al. 2022.
\newblock UnifiedSKG: Unifying and Multi-Tasking Structured Knowledge Grounding
  with Text-to-Text Language Models.
\newblock \emph{arXiv preprint arXiv:2201.05966}.

\bibitem[{Xu et~al.(2021)Xu, Kumar, Yang, Zi, Tang, Huang, Cheung, Prince, and
  Cao}]{dtfixup2021xu}
Xu, P.; Kumar, D.; Yang, W.; Zi, W.; Tang, K.; Huang, C.; Cheung, J. C.~K.;
  Prince, S.~J.; and Cao, Y. 2021.
\newblock Optimizing Deeper Transformers on Small Datasets.
\newblock In \emph{Proceedings of the 59th Annual Meeting of the Association
  for Computational Linguistics and the 11th International Joint Conference on
  Natural Language Processing (Volume 1: Long Papers)}, 2089--2102. Online:
  Association for Computational Linguistics.

\bibitem[{Yu et~al.(2020)Yu, Wu, Lin, Tan, Yang, Radev, Xiong
  et~al.}]{grappa2020yu}
Yu, T.; Wu, C.-S.; Lin, X.~V.; Tan, Y.~C.; Yang, X.; Radev, D.; Xiong, C.;
  et~al. 2020.
\newblock GraPPa: Grammar-Augmented Pre-Training for Table Semantic Parsing.
\newblock In \emph{International Conference on Learning Representations}.

\bibitem[{Yu et~al.(2018)Yu, Zhang, Yang, Yasunaga, Wang, Li, Ma, Li, Yao,
  Roman et~al.}]{spider2018yu}
Yu, T.; Zhang, R.; Yang, K.; Yasunaga, M.; Wang, D.; Li, Z.; Ma, J.; Li, I.;
  Yao, Q.; Roman, S.; et~al. 2018.
\newblock Spider: A Large-Scale Human-Labeled Dataset for Complex and
  Cross-Domain Semantic Parsing and Text-to-SQL Task.
\newblock In \emph{Proceedings of the 2018 Conference on Empirical Methods in
  Natural Language Processing}, 3911--3921.

\bibitem[{Zelle and Mooney(1996)}]{data-geography-original}
Zelle, J.~M.; and Mooney, R.~J. 1996.
\newblock Learning to Parse Database Queries using Inductive Logic Programming.
\newblock In \emph{AAAI/IAAI}, 1050--1055. Portland, OR: AAAI Press/MIT Press.

\bibitem[{Zhao et~al.(2021)Zhao, Wallace, Feng, Klein, and
  Singh}]{pmlr-v139-zhao21c}
Zhao, Z.; Wallace, E.; Feng, S.; Klein, D.; and Singh, S. 2021.
\newblock Calibrate Before Use: Improving Few-shot Performance of Language
  Models.
\newblock In Meila, M.; and Zhang, T., eds., \emph{Proceedings of the 38th
  International Conference on Machine Learning}, volume 139 of
  \emph{Proceedings of Machine Learning Research}, 12697--12706. PMLR.

\end{thebibliography}
